\documentclass[letterpaper, 10 pt, conference]{ieeeconf}  

\IEEEoverridecommandlockouts                              

\usepackage{amsmath}
\usepackage{amssymb}
\usepackage{graphicx}
\usepackage{cite}
\usepackage{booktabs}

\usepackage{amsthm}

\usepackage[dvipsnames]{xcolor}
\usepackage{subcaption}
\usepackage{wrapfig}
\usepackage{multirow}
\usepackage{mathtools}
\usepackage{empheq}
\usepackage{color}
\usepackage{bbm}
\usepackage{xspace}
\let\labelindent\relax
\usepackage[shortlabels]{enumitem}

\usepackage{algorithmicx}
\usepackage{algorithm}
\usepackage[]{algpseudocode}

\usepackage{fontawesome}

\usepackage{soul}
\usepackage{comment}

\usepackage[hyphens]{url}
\usepackage[pdftex, colorlinks=true, linkcolor=blue, citecolor = black, urlcolor = black, filecolor=black , pagebackref=true, hypertexnames=false]{hyperref}
\usepackage{cleveref}
\crefname{section}{Sec.}{Secs.}
\Crefname{section}{Section}{Sections}
\Crefname{table}{Table}{Tables}
\crefname{table}{Tab.}{Tabs.}
\crefname{figure}{Fig.}{Figs.}

\newcommand{\bmat}{\begin{bmatrix}}
\newcommand{\emat}{\end{bmatrix}}

\title{\LARGE \bf
DC-MRTA: Decentralized Multi-Robot Task Allocation and Navigation in Complex Environments}

\author{Aakriti Agrawal$^{1}$, Senthil Hariharan$^{1}$, Amrit Singh Bedi$^{1}$,  Dinesh Manocha$^{1}$ 
\thanks{$^{1}$University of Maryland, College Park, MD, USA {\tt\small \{agrawal5,sarul1,amritbd,dmanocha\}@umd.edu}}%
\thanks{$^{2}${This work was supported in part by ARO Grants W911NF1910069, W911NF2110026  and U.S. Army Cooperative Agreement W911NF2120076}
        }%
}

\begin{document}

\maketitle

\begin{abstract}
We present  a novel reinforcement learning (RL) based task allocation and decentralized navigation algorithm for mobile robots in warehouse environments. Our approach is designed for scenarios in which multiple robots are  used to perform various pick up and delivery tasks. We consider the problem of joint decentralized task allocation and navigation and present a two level approach to solve it. At the higher level, we solve the task allocation by formulating it in terms of Markov Decision Processes and choosing the appropriate rewards to minimize the Total Travel Delay (TTD). At the lower level, we use a decentralized navigation scheme based on ORCA that enables each robot to perform these tasks in an independent manner, and avoid collisions with other robots and dynamic obstacles. We combine these lower and upper levels by defining rewards for the higher level as the feedback from the lower level navigation algorithm. We perform extensive evaluation in complex warehouse layouts with large number of agents and highlight the benefits over state-of-the-art algorithms based on myopic pickup distance minimization and regret-based task selection. We observe improvement up to $14\%$ in terms of task completion time and up-to $40\%$ improvement in terms of computing collision-free trajectories for the robots.

\end{abstract}

    







	

%
%
\section{Introduction}
Multiple robots are ubiquitous and are used to perform various tasks in different applications such as planetary exploration \cite{planetary}, search and rescue missions \cite{searchandrescue}, and automated manufacturing \cite{automated_manufacturing}. Recently, there has been considerable interest in using robots for warehouse automation.  These environments are based on large-scale fulfillment centers that organize inventory using several movable pods. The pods get transported by a team of robots to different picking stations, where each item gets picked by a human operator~\cite{pick_up_and_delivery,warehouse_comb}.

A key issue is designing a multi-robot decision making system that can improve the overall throughput and result in safe navigation of robots. The overall problem consists of two challenges for each robot: \emph{task allocation} and \emph{navigation}. The problem is challenging because these two parts are coupled with each other.  The task allocation controls the starting and destination of the navigation part, and the navigation part decides the next tasks to be allocated. As large numbers of robots are deployed, another  key challenge is collision-free path computation.  Not only does each robot needs to avoid collisions with static obstacles and other moving robots in the scene, but it must also avoid any dynamic obstacles or humans moving in the environment. 

In the existing literature, the problems of multi robot task allocation and multi robot navigation are assumed to be decoupled and solved separately~\cite{mapf_decoupled}. The tasks are typically allocated based on greedy and auction based approaches~\cite{market_effi,market_robust}. The navigation paths can be computed via different centralized or decentralized algorithm~\cite{astar,orca,sipp,cbs,vrvo,Luna2011PushAS}. In centralized  navigation systems~\cite{cbs,sipp,MILP,MIQP}, a single controller plans the path of all agents and instructs the agent in real time during execution. Agents, also need to provide feedback to the central controller in real time.   This approach is regarded as more stable, provides better guarantees in terms of reaching the goal, avoids deadlock, and can handle narrow passages in the environment. However, a centralized system
requires major investment in terms of a central control system, a communication system and a highly accurate navigation system. Furthermore, it creates a bottleneck~\cite{velagapudi2010decentralized}. A failure in terms of centralized computation could lead to complete system failure or small errors or miscommunication can lead to accidents, since the robots cannot make independent collision-free navigation decisions.
 Furthermore, these centralized techniques may not be able to handle unstructured environments with human co-workers or unknown obstacles.
Some of the MAPF methods used in such warehouses 
generate piece-wise linear paths that involve ``stop and turn'' maneuvers, and this may not work well for non-holonomic agents~\cite{cohen2019optimal}. 

In this work, our goal is to develop efficient methods that can utilize decentralized navigation schemes~\cite{orca}, as they can handle dynamic obstacles and unknown environments. These decentralized methods can scale to a large number of agents without using a centralized controller and perform local navigation to compute collision-free trajectories.  Furthermore, they can be used for robots with car-like or non-holonomic constraints~\cite{wilkie2009generalized,alonso2013optimal}. Moreover, we would like to treat these two problems of \emph{task allocation} and \emph{navigation} as part of an overall coupled warehouse automation system~\cite{usc} to improve the overall efficiency in terms of task completion time and navigation metrics.

\textbf{Main Results:}
We present a novel and coupled solution for multi-robot task allocation and decentralized navigation (DC-MRTA). In our coupled task-allocation formulation, tasks are allocated to the agents simultaneously during navigation.  We use a reinforcement learning (RL)-based strategy that can be combined with state-of-the-art decentralized MAPF methods~\cite{orca}.
Our goal is to assign tasks in a manner that the total time taken to do nothing called \em{total travel time (TTD)} is minimized. It also further improves the \em{(makespan)}. Our approach is general and does not make any assumptions about the environment or the number of robots or the tasks.
The main novel results include:

\begin{itemize}
\setlength\itemsep{0.25em}
    \item  We propose a novel two level approach for the problem of joint decentralized task allocation and navigation in warehouse environments.  We formulate the problem of decentralized multi-robot task allocation as a Markov Decision Processes and use a reinforcement learning (RL) based strategy to solve it, which can be combined with any decentralized navigation method. 

    \item We present a novel coupled architecture to solve the problem in a warehouse environment. Our architecture can handle variable number of robots and tasks, which is particularly important in decentralized navigation settings. 
    
    \item The rewards for the high level RL method are defined based on the feedback from the low-level navigation algorithm. In particular, we use total travel delay (TTD), which is generated from the decentralized navigation algorithm, as a reward to train the policy for task allocation. The trained policy can be easily combined with other decentralized methods like ORCA.

\end{itemize} 
  We perform extensive evaluations and compare against the state-of-the-art greedy baselines myopic pickup distance minimization (MPDM) and regret-based task-selection (RBTS). We test performance on different warehouse layouts with varying number of robots and show an improvement of up to $14\%$ in terms of task completion time for $500$ tasks. Our scales well with the number of agents and can handle thousands of agents. For dense scenarios, our approach also reduces the number of inter-agent collisions by $40\%$.

\section{Related Work}

\subsection{Multi-Robot Decentralized Navigation} 
Multi-robot navigation is a widely researched topic in robotics and related areas that involves the computation of collision-free paths for robots in a multi-agent setting. Centralized and decentralized methods are two categories of navigation algorithms that differ based on the decision-making entity and available knowledge of its environment.

Our approach is based on  decentralized planners, where the agents use the knowledge of their local environment to make independent navigation decisions. The agent's local information includes the position (and velocity) and the dimension information of their neighbors. Velocity Obstacles (VO)~\cite{VO} presents an influential class of decentralized algorithms that compute a set of velocities that could result in a collision. RVO~\cite{RVO} improve upon VO by accounting for the agent's reactive behavior, thus reducing oscillations in trajectories. RVO was further posed as a linear convex optimization in ORCA~\cite{orca} and extended to different agent dynamics~\cite{AVO,NH-ORCA}. Buffered Voronoi Cell (BVC)~\cite{BVC} constructs retracted Voronoi cells to limit the agent's local motion to free space. In contrast to VO-based methods, BVC only uses the position information of its neighbors. Decentralized methods can be scaled to large number of agents, however it is difficult to guarantee optimality or deadlock-free navigation especially in obstacle rich environments or narrow passages. More recently, RL-based methods~\cite{fan2020distributed,GA3C_CADRL,tan2020deepmnavigate} can also be used to improve the collision avoidance performance and compute shorter time to goal. Our approach is general and can be combined with any of these decentralized schemes.

\subsection{Multi-Robot Task Allocation (MRTA)} 

MRTA involves the opti9mal assignment of robots to tasks. It is a variant of the multiple Traveling Salesman Problem (mTSP), which is known to be NP-hard to solve optimally~\cite{yousefikhoshbakht2013modification,6721865}. Extensive surveys are available in~\cite{taxonomy,Khamis2015,Tkach2020}. Warehouse task allocation is studied as integer linear program in~\cite{warehouse_comb}. In~\cite{warehouse_hetro_robot}, an optimal task allocation algorithm is proposed for  heterogeneous agents. These methods, optimisation based, assume all agents are available at once for optimal matching. 

Thus are two widely used approaches for solving these problems: (1) market-based approaches and (2) optimization-based approaches for different environments~\cite{badreldin2013comparative}. Auction-based approaches\cite{brunet2008consensus} (CBBA, CBAA), which are part of market-based approaches, have also been proposed for decentralized task allocation. These algorithms use local information to assign tasks and consensus to improve global objective. Variants of CBBA have been proposed \cite{choudhury2021dynamic} like deep-RL based communication aware variants ~\cite{CA-CBBA} for environments with limited bandwidth. We use one regret based auction method in our baseline comparisons. 

There is some work~\cite{ames_decTask,ghassemi2018decentralized,liu2013optimal}  on decentralized task-allocation and navigation. \cite{ames_decTask} is a RL based method coupled with navigation but does not show scalability. \cite{ghassemi2018decentralized} proposes a decentralised task-robot matching method based on bipartite graph. It performs task-clustering followed by maximal matching for task allocation. Overall its performance is about   $7-28\%$ of the optimal cost obtained using centralized schemes. 
These decentralised methods have not been tested along with decentralised navigation algorithms, complex environments and large number of robots and tasks.

\begin{figure*}[!htb]
    \centering
    \begin{minipage}{\textwidth}
        \centering
       \hspace{0cm}\includegraphics[scale=0.2]{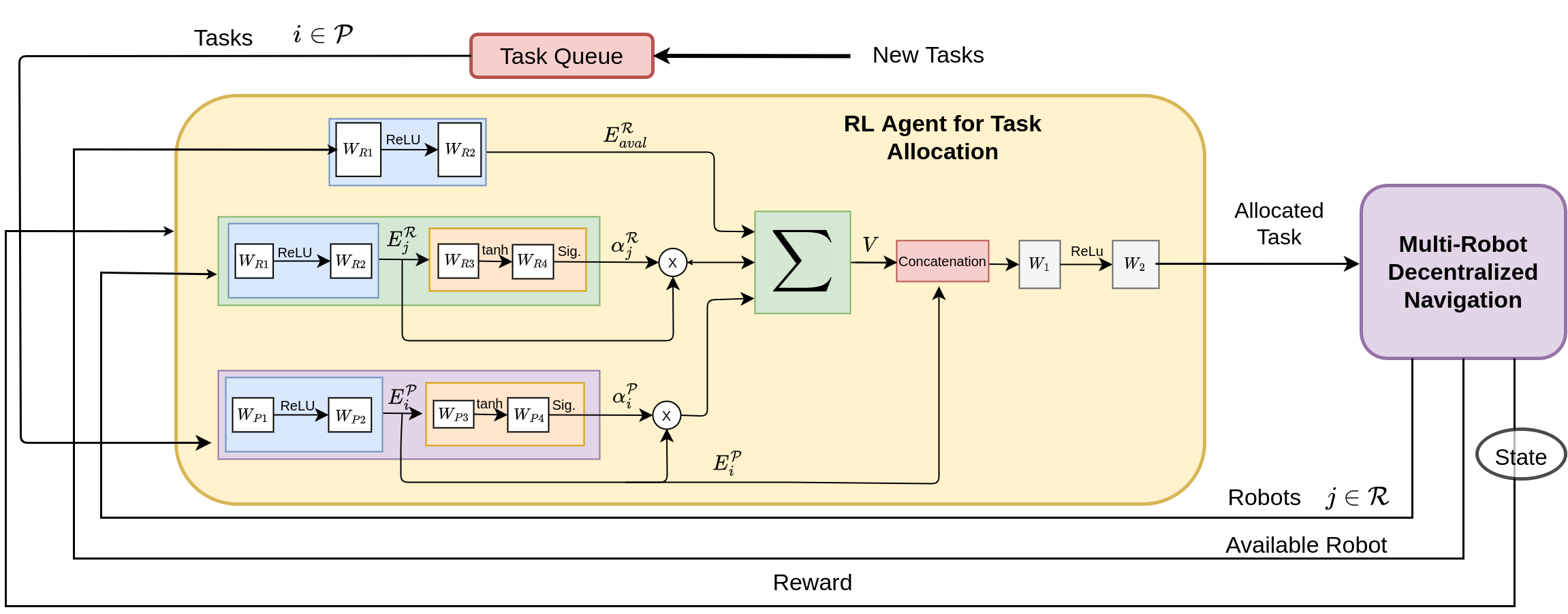}
\caption{Task Allocation and Decentralized Navigation: We highlight our approach which uses RL-based task allocation followed by ORCA-based decentralized navigation.  We highlight the coupling between the high level RL algorithm and the low level navigation method. Our task allocation method uses heuristics that tend to reduce congestion in terms of multi-agent navigation, while the feedback from collision avoidance is used to formulate the reward functions for RL agents. Our proposed  RL agent is able to deal with variable number of tasks and agents shown by the weighted averaging block to form global task and agent embeddings. }        \label{fig:network}
\vspace*{-0.1in}
    \end{minipage}%
\end{figure*}

\section{Background and Problem Formulation}\label{formulation}
In this section, we give an overview of our problem of task allocation and navigation in complex, warehouse environments.

\subsection{Decentralized Multi-Robot Task Allocation and Navigation}

We consider a group of $n$ holonomic, disk shaped robots operating in an environment $\mathcal{W} \subset \mathbb{R}^2$. For any agent $A_i$, its geometry is represented by $\mathcal{A}_i$. The agent's position is given by $\mathbf{p}_i$, and we consider our agents to be velocity ($\mathbf{v}_i$) controlled. Agents within a defined sensing radius of agent $i$ are considered its neighbors, and are represented by the set $\mathcal{N}_i$. We formulate a decentralized navigation problem where each robot makes independent decision to perform these tasks and avoid collisions with other agents and static/dynamic obstacles using the local environment knowledge.  We assume each agent $A_i$ knows its neighbors' positions, where neighbors are agents located within a certain radius of $A_i$. This information can be obtained using perception or communication modules. Overall,
Agent $A_i$ local knowledge includes $\mathbf{p}_j, \mathbf{v}_j \quad j \in \mathcal{N}_i $.  Similarly, it has a representation of all the obstacles in the environment. We use the standard A* method to compute a path for each agent that avoids collisions with all the static obstacles.

A task is represented by a tuple $(\mathbf{o_i}, \mathbf{d_i})$ which includes the origin $\mathbf{o_i}$ and destination $\mathbf{d_i}$ locations in the environment. The goal is to allocate a task to available agent $i$ effectively. After the task assignment, the agent's goal is set as $\mathcal{G}_i = \{\mathbf{o_i}, \mathbf{d_i}\}$, as it represents the pickup and drop location. This is an important difference here from the standard decentralized navigation problems where there is no dependence of goal on another task allocation module. Instead the time taken by the navigation algorithm to compute a path also affects the performance of the task allocation scheme.

\subsection{Metrics}
Our goal is to design a scheme where we consider single-task robots which execute only a single task at any given time. We assume that all the tasks are single robot tasks  and task assignment is instantaneous, where the task is instantaneously allocated at each decision timestep. Our approach is designed in terms of a sequential-lifelong solution,  where new tasks are continuously generated and task is allocated to a robot as soon as it becomes available. We would like to maximize the sum of utilities over time, that is, assign tasks at each instant in such a way that the total time taken to execute a specific number of tasks \em{makespan} as well as \em{TTD} is minimized. TTD is defined as the time taken to travel from the current location of the robot $\mathbf{p}_i$ to the starting location of the allocated task $\mathbf{o}_i$.  

Given a specific task, our next step is to navigate the robot from their current position $\mathbf{p}_i$ to $\mathbf{o_i}$, and then to $\mathbf{d_i}$, while remaining collision-free. At any timestep, an agent $A_i$ is said to be collision-free provided its geometry ($\mathcal{A}_i$) does not overlap with its neighbors or static or dynamic obstacles. Assuming $m$ obstacles in the environment, their geometries can be represented by $\mathcal{O}_j, \quad j \in {1,2,...,m}$. The agent sequentially reaches the goals in the goal set ($\mathcal{G}$) while {\em avoiding collisions}. That is, at each time step the agent moves towards its immediate goal $( \mathbf{g_{i}} \in \mathcal{G}_i = \{\mathbf{o_i}, \mathbf{d_i}\})$. 
Our goal is to design navigation methods that tend to reduce or eliminate any collisions.

\begin{equation}
\begin{aligned}
\min \Vert \mathbf{g}_{i} -& \mathbf{p}_i \Vert \\
\mathcal{A}_i \cap \mathcal{A}_j &= \emptyset, \quad \forall j \in \mathcal{N}_i \\
\mathcal{A}_i \cap \mathcal{O}_j &= \emptyset, \quad \forall j \in \{1,2,...,m\}. 
\label{problem}
\end{aligned}
\end{equation}

On reaching the destination location ($\mathbf{d_i}$), the task is considered to be complete and the agent is once again available to be allocated a task. We note that in problem \eqref{problem}, the objective $\Vert \mathbf{g}_{i} - \mathbf{p}_i \Vert$ is not fixed and depends upon the task allocation. Therefore, jointly solving the problem for optimal $\mathbf{g}_i$ at each $i$, and then designing robot trajectories is difficult in practice.

\section{DC-MRTA: Our Two Level Coupled Approach}
In this section, we present our approach to solve the joint problem in Equation \eqref{problem} using a two level scheme. Our proposed solution is summarized in Fig.~\ref{fig:network} and algorithm \ref{alg:phedocode}. On the higher level, we design a reinforcement learning based task allocation strategy, and on the lower level, a multi robot decentralized navigation algorithm (ORCA) is used. We describe both the levels next in detail. 

\subsection{Lower level:Decentralized  Navigation}\label{lower}
We first consider the case when $\mathbf{g_i}$ fixed. We consider simple navigation schemes based on A* and also  use Optimal Reciprocal Collision Avoidance (ORCA) algorithm to move the robot to that goal location with no inter-agent collisions. The specification of $\mathbf{g_i}$ actually comes from the higher level task allocation algorithm. 
Given $n$ homogeneous, disk-shaped agents with position $\mathbf{p}_i$, velocity $\mathbf{v}_i$, and radius $r_i$, ORCA computes a suitable collision-free velocity to drive each agent towards their respective goal location $\mathbf{g}_i$, while avoiding collisions with other obstacles and agents. 

For any two agents $i, j$, the $VO^{\tau}_{i \vert j}$ is a set of relative velocities that can lead to collision within the time horizon $\tau$. Geometrically, it is represented by
\begin{equation}
VO^{\tau}_{i\vert j} = \{ \mathbf{v} \vert \forall t \in [0,\tau] , t\mathbf{v} \in D(\mathbf{p}_j - \mathbf{p}_i, {r}_i + {r}_j) \}, \label{orca}
\end{equation}
where, $D(\mathbf{p},r)$ represents a disk with center $\mathbf{p}$ and radius $r$. 
Assuming the agent uses the  velocities $\mathbf{v}^{opt}_i$ and $\mathbf{v}^{opt}_j$, the agents are collision bound if $\mathbf{v}^{opt}_i - \mathbf{v}^{opt}_j \in VO^{\tau}_{i \vert j}$. Let $\mathbf{u}$ be the closest point on the boundary of $VO^{\tau}_{i \vert j}$. Then, $\mathbf{u}$ provides the minimum change in relative velocity to avoid collision. Taking $\mathbf{n}$ as the outward normal at $ (\mathbf{v}^{opt}_i - \mathbf{v}^{opt}_j) + \mathbf{u}$, the set of collision-free velocities for agent $i$ can be represented geometrically as a half-plane, given by
$$
ORCA^{\tau}_{i \vert j} = \{ \mathbf{v} \vert (\mathbf{v} - (\mathbf{v}^{opt}_i + \frac{1}{2} \mathbf{u})) \cdot \mathbf{n} \ge 0\}.
$$
A suitable velocity is chosen from the set $ORCA^{\tau}_{i \vert j}$ by minimizing the distance with $\mathbf{v}^{pref}_i$ which is the agent's preferred velocity directed towards the goal.  

 ORCA uses these conservative constraints to provide collision avoidance guarantees~\cite{orca}. For each neighboring agent, the ORCA constraint is represented as a half-plane (Equation \ref{orca}). If there is a feasible solution to the linear programming formulation, that provides a sufficient condition. However, there may be no feasible solution, especially if there are  dense scenarios with a large number of agents in close proximity or narrow passages. In other words, scenarios with high level of congestion can result in possible collisions. Our goal is to design a task-allocation scheme such that it tends to minimize congestion. As a result, we need to develop methods that can:
\begin{itemize}
    \item The task allocation procedure needs to  choose appropriate task positions, ($\mathbf{d_i}$), for a robot such that it is not in close proximity to the other agents or result in congested scenarios. Another goal is to avoid deadlock scenarios where some agents can block other scenarios or multiple agents are in narrow passages.
    \item The decentralized navigation scheme uses local information for collision avoidance with the agents and obstacles. The underlying scheme used to modify the velocity of each agent to avoid collisions is used to design the appropriate reward functions for our RL methods.
    \end{itemize}

\subsection{Higher Level: RL based Task Allocation}
The problem of multi-robot task allocation is mainly solved via optimization methods in the existing literature \cite{task_survey}. Such methods won't work in our setting because we are working with decentralized navigation schemes and account for the criteria highlighted above. Since robots behave independently in the warehouse environment, they become available at different time instances. This makes it hard to solve with existing optimization based approaches, especially we also take into account the constraints of the navigation scheme. Additionally, the number of robots and tasks is also not fixed a priori which further makes the problem more challenging. To address these issues, we formulate the problem as a reinforcement leaning based task allocation in warehouse environments.
The RL formulation is non-trivial because it requires defining a Markov Decision Process (MDP) which includes the design of state spaces, action spaces, and rewards etc. We proceed next to describe this MDP in detail. 

An MDP here is defined by a tuple  $\mathcal{M}:=\left(\mathcal{S}, \mathcal{A}, R, {P}, \gamma\right)$, where state space $\mathcal{S}$, action space $\mathcal{A}$, and $R(s,a)$ denotes the reward. Here ${P}$ describes the transition probability matrix where ${P_a(s,s')}$ is the probability of transition to state $s'\in\mathcal{S}$ from state $s\in\mathcal{S}$ after taking action $a\in\mathcal{A}$.  In MDP, $\gamma\in(0,1)$ denotes the discount factor. To solve the problem, we need to explicitly define the terms in MDP $\mathcal{M}$.
We denote time instance as $t\in\mathbb{N}$, $M$ as number of robots given by $\mathcal{R}=\{1,2,\cdots,M\}$. The task allocation flow is described as follows. 

After the task allocation at $t$, we remove the allocated task from the queue and a new task is added. Since we are looking at the joint task allocation and navigation problem, we assume for tractability that only one robot is available at any instance $t$. We take one step further from $t$ to $t+1$ as soon as a robot becomes available. Next, we define in detail the state space $\mathcal{S}$, action space $\mathcal{A}$, and reward $R$ for the warehouse environment.

\textbf{State:} Each task $i\in \mathcal{P}$ is mathematically represented by a tuple $(\mathbf{o}_i, \mathbf{d}_i, k_i, l_i)$, where $\mathbf{o}_i$ is the origin of the task, $\mathbf{d}_i$ is the destination, $k_i$ denotes the distance between the selected robot's current position to $\mathbf{o}_i$, and $l_i$ is the length (distance between $\mathbf{o}_i$ and $\mathbf{d}_i$) of the task.  Each robot $j\in\mathcal{R}$ is also represented by a tuple $(\mathbf{p}_j, r_j)$, where $p_j$ is position ($X-Y$ coordinates) in the $2$D map and $r_j$ is the time left to complete the allocated task (i.e., the task completion time). {Let $j_{sel}$ denotes the available robot for task allocation.} Then, the state $s$ encapsulates the robots' positions, their task completion times, and the task list  $\mathcal{P}$, which consists of tuples $(\mathbf{o}_i, \mathbf{d}_i, k_i, l_i)$ for all the tasks and the selected robot {$j_{sel}$}. Collectively, we write the state $s:=\big\{(\mathbf{p}_j, r_j)_{\forall j\in \mathcal{R}}, (\mathbf{o}_i, \mathbf{d}_i, k_i, l_i)_{\forall i\in \mathcal{P}}, j_{sel} \big\}$ and the collection of all possible states $s$ constitutes the state space $\mathcal{S}$.

\textbf{Action:} An action $a$ in the environment is used to select which task to execute from the list of $N$ available tasks in the queue. We define a policy $\pi:\mathcal{S}\rightarrow \mathcal{D}(\mathcal{A})$ (here $\mathcal{D}$ denotes the set of possible distributions defined over $\mathcal{A}$) which takes the current state as input and outputs a distribution across the all possible actions $a\in\mathcal{A}$. Note that in our formulation, $\mathcal{A}$ constitutes the list of all available tasks from which we need to select a particular task for the selected robot $j_{sel}$. During the training time, we select the action $a\sim \pi(\cdots ~|~ s)$ where  $\pi(\cdot ~|~ s)$ is the probability distribution across the tasks. During the test time, we select the action in a deterministic manner  by choosing a task to accomplish with maximum probability in $\pi(\cdot ~|~ s)$. Further, it is important to choose the  specific policy architecture for our approach.
In the literature, common choices for policy is to define a deep neural network (DNN) \cite{DBLP:journals/corr/abs-1909-01150} denoted as $\pi_{\theta}(\cdot ~|~ s)$, where $\theta$  corresponds to neural network weights. The design of policy architecture is important because it actually defines the space of policies over which we search for the optimal actions. One could try a simple NN architecture where we pass all the available robot information (locations, availability) to the network and train it. But unfortunately, such simple designs doesn't result in convergence and also not scalable with respect to number of robots and tasks. Hence, we carefully designed the architecture using attention based mechanisms \cite{attention} to make sure that policy network supports variable number of robots and tasks.

\subsection{Coupling via Reward Design}
Designing the reward signal for any RL problem is the most crucial part. This is because the reward function decides the behavior of the policy we are interested in. We could design the reward in multiple ways which would eventually decide whether there is any coupling between the lower level (navigation) and higher level (task allocation). One simple way to design a reward for the task allocation would be to just consider $0/1$  reward (reward is zero if the task is accomplished, otherwise 1). This reward makes sense, but it completely decouples the task allocation from navigation algorithm trajectories. For instance, this reward would  be agnostic to any collision which occurs during the robot navigation, as highlighted in Section IV(A).

In our approach, we use a more intelligent way to design rewards called \emph{coupled rewards,} which utilizes information from navigation algorithm to design efficient task allocation policy. For instance, we define our reward as 
\begin{align}
    \text{reward}= -1\left(\text{Time}(\mathbf{o}_i,\mathbf{p}_i)\right),
\end{align}
where $\mathbf{p}_i$ is the current location of the robot, and $\left(\text{Time}(\mathbf{o}_i,\mathbf{p}_i)\right)$ evaluates the time a robot takes to travel from $\mathbf{p}_i$ to $\mathbf{o}_i$. This is also known as Total Travel Delay (TTD) and we are interested in minimizing it. This is evaluated by navigating the robot in the environment via a lower level decentralized navigation algorithm and hence creates a coupling between lower and upper levels.

\textbf{State-Transitions:} At current state $s_t$, we have a task list in queue $\mathcal{P}$ and one available robot (say $i\in\mathcal{R}$). At this state, we take action $a_t$ according to a policy $\pi(\cdot~|~s_t)$ that decides which task would be allocated to the free robot. The selected task is removed from the list, new task gets added to the list, and the robot becomes unavailable. Our algorithm will wait until next robot becomes available which would mark our new state $s_{t+1}$. Note that the state transitions do not happen at regular time intervals here.

\begin{algorithm}
\begin{algorithmic}

\caption{Decentralised Task-Allocation and Navigation}

\label{alg:phedocode}
\State Let $s:=\big\{(\mathbf{p}_j, r_j)_{\forall j\in \mathcal{R}}, (\mathbf{o}_i, \mathbf{d}_i, k_i, l_i)_{\forall i\in \mathcal{P}}, j_{sel} \big\}$ be current state of the system.
\State Each robot $j\in\mathcal{R}$ is executing a task  $(\mathbf{o}_j, \mathbf{d}_j)$
\State Robot's current position is denoted as $\mathbf{p}_j$.
\State Let time be $t$
\While{True} 
\State Update $\mathbf{p}_j$ for all $j\in\mathcal{R}$ by unit amount using ORCA. 
\State $t \leftarrow t+1$

\If {Any $\mathbf{p}_j$ $==$ $\mathbf{d}_j$}
\State $j_{sel} \leftarrow j$
\State $(\mathbf{o}_{j_{sel}}, \mathbf{d}_{j_{sel}})  \leftarrow \text{MRTA} (j_{sel}, s)$ 

\EndIf
\EndWhile
\Statex
\Function{MRTA}{$j_{sel}$, $s$} 
    \State RL-policy takes state $s$ as input and selects a task $i\in \mathcal{P}$ as action.
    \State $(\mathbf{o}_{j_{sel}}, \mathbf{d}_{j_{sel}}) \leftarrow (\mathbf{o}_{i}, \mathbf{d}_{i})$
    \State \Return $(\mathbf{o}_{j_{sel}}, \mathbf{d}_{j_{sel}})$
\EndFunction

\end{algorithmic}

\end{algorithm}

\section{Implementation and Results}\label{experiments}

In this section, we describe the implementation of our network architecture and analyze its performance in different settings by varying the number of robots, task generation schemes, layouts and navigation schemes. During training, we have used two different planning or navigation techniques to compute the reward : \emph{direct navigation} and \emph{A*}. In the direct navigation approach, the Euclidean distance between the robot position and task origin is used to evaluate the cost at each instant $t$. Our A* navigation scheme takes into account the static obstacles in a scene and  computes collision-free paths for the agents. As mentioned earlier Sec. IV(C), the negative of this calculated cost is defined as the reward in the environment and it also serves as a heuristic to reduce congestion.

\vspace{-2mm}
\subsection{A*+ORCA:Decentralized Navigation}
\label{Astar_ORCA}
\vspace{-0.5mm}
 Our task allocation  runs on model trained only on A* for a specific layout as that accounts for the static obstacles and uses that information to choose the appropriate rewards. During testing, we also evaluated our method using a combination of  A* and ORCA decentralised navigation algorithms along with \emph{direct navigation} and \emph{A* navigation} algorithms. In this case, the ORCA formulation tries to avoid collisions between the moving agents.
 In order to travel through static obstacles, we need a global navigation algorithm which can generate trajectory points between start and goal points. We use single agent A*\cite{astar} to find the shortest path between two locations that avoids the obstacles. That path is used as the initial guidance (or preferred velocity) for ORCA agent. It is modified to avoid collisions with other agents or dynamic obstacles.
 
The default parameters used in our implementation are radius is 1.5 unit, time-step is 0.25 seconds, max velocity is 2 units/sec.






\subsection{Baselines}
We compare our algorithm, DC-MRTA, with minimum pickup distance minimization (MPDM) and regret based task selection (RBTS) as our baselines. Both these baselines are used in prior literature.

\subsubsection{MPDM} algorithm chooses the task that is closest to the robot. It is optimal if there is only one robot in the system and the task utility equals the distance of the robot to the task start position~\cite{taxi_comb_opti}.  MPDM is the one of the widely used baseline model used for comparison in the literature for decoupled task allocation and navigation~\cite{mapf_decoupled}. 

\subsubsection{RBTS}
We use a modified version of RBTS used in \cite{usc}. It is inspired from \cite{rbts_1,rbts_2}. For every task in the list, find the distance to the closest robot. Subtract the distance of the selected robot to each of these tasks with the previous quantity. Choose the task with the maximum value (i.e., the maximum regret). 

For all our benchmarks, we run the test on a fixed task set by fixing the seed of the random number generator and report TTD values for them. We train our model on the  Nvidia GeForce RTX $2080 Ti$ GPU with $11$GB of memory, for $5$ different environments and batch size of $32$. We run the training for $4$ million iterations, which takes around $12$ hours with our A* navigation algorithm. The network architecture is large and GPU usage during training is usually less than 5-10 \%. Next, we describe the different experimental settings and results.

\begin{figure}[H]
	\centering
	\begin{subfigure}{.2\columnwidth}
		\includegraphics[trim={0.3cm 0.2cm 0.5cm 0.4cm},scale=0.26]{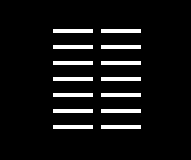}
		\caption{Layout A.}
	\end{subfigure}%
	\hspace{4mm}
	\begin{subfigure}{.2\columnwidth}
		\includegraphics[trim={0.8cm 0.8cm 0.8cm 1cm},scale=0.25]{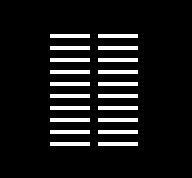}
		\caption{Layout B.}
	\end{subfigure}
		\hspace{2mm}
	\begin{subfigure}{.2\columnwidth}
		\includegraphics[trim={0.7cm 0.8cm 0.7cm 1.0cm},scale=0.26]{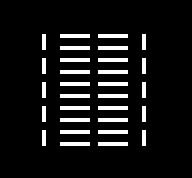}
		\caption{ Layout C.}
	\end{subfigure}%
	%
	\vspace{2mm}

	\begin{subfigure}{.2\columnwidth}
		\includegraphics[trim={0.6cm 0.6cm 0.6cm 0.4cm},scale=0.26]{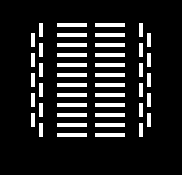}
		\caption{ Layout D.}
	\end{subfigure}
	\begin{subfigure}{.2\columnwidth}
		\includegraphics[trim={0.6cm 0.6cm 0.6cm 0.4cm},scale=0.26]{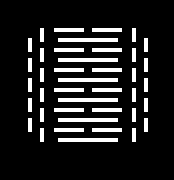}
		\caption{Layout E.}
	\end{subfigure}
	\caption{ We highlight five different warehouse environment's layout which we consider for evaluation. They have many obstacles (shown in white) and narrow gaps between them. Many simple task allocation and navigation methods do not work well, when we consider a large number of agents. 
}
	\label{fig:layout_other}
	\vspace*{-0.1in}
\end{figure}

\section{Evaluation}

\subsection{Experimental Setting}
We generate tasks randomly in the environment using a designated method as explained below. We further explain the layout generation in detail, which we use for testing the proposed algorithm. 

\textbf{Task and Layout  Generation:} To the best of our knowledge, real data sets are not available for the problem of multi-robot task allocation and navigation. Mostly in literature, a synthetic procedure is used to generate data and then test the proposed algorithms \cite{mapd_review}. In this work, we utilize the designated method for task generation (see \cite{mapd_review} for further details on this method). The idea is to define different regions for pickup and delivery tasks and then generate random samples from them \cite{task_gen,task_gen2}.  For the layout generation, the authors in \cite{mapd_review} describes publically available warehouse layouts generated through Asprilo. Asprilo is an open source framework used to simulate automated warehouse scenarios \cite{mapd_review}. We generate different layouts with varying the level of compactness for a specific size (see Fig. \ref{fig:layout_other}).

\subsection{Simple Navigation Schemes}
We highlight the results in Table \ref{tab:simple_time} for direct navigation and A* navigation, respectively. We use the layouts as shown in Fig~\ref{fig:layout_other} for this evaluation. In this setting, training and testing is performed on the same navigation algorithm i.e. direct and A*.

\begin{table}[t]
	\centering
	\resizebox{1.0\columnwidth}{!}{%
	\begin{tabular}{|c|c|c|c|c|c|c|} 
		\hline
		\multicolumn{1}{|l|}{}             & Robot No. & MPDM  & RBTS & DC-MRTA & \% Imp. MPDM & \% Imp. RBTS \\
		\hline
		Direct & 10            & 7285.2  & 7820 & \textbf{6984.3} & 4.13 & 10.68  \\
		\cline{2-7}
		& 100           & 12151.5  & 12048 & \textbf{11420.1} & 6.01 & 5.21   \\ 
		\hline
		Layout  & 10 & 6332 & 6233 & \textbf{6208} & 1.95 & 0.40  \\  
		\cline{2-7}
		 A & 100 & 6836 & 7195 & \textbf{6616} & 3.21 & 8.04  \\  
		\hline
		Layout  & 10            & 5989 & 6351 & \textbf{5732} & 4.29 & 9.74 \\  
		\cline{2-7}
		 B & 100           & 7472 & 7609 & \textbf{7112} & 4.81 & 6.53 \\  
		\hline
		Layout  & 10            & 5992 & 6204 & \textbf{5804} & 3.13 & 6.44 \\  
		\cline{2-7}
		 C & 100           &  7511 & 7664 & \textbf{7403} & 1.43 & 3.40  \\  
		\hline
		Layout  & 10  & 6157 & 6153 & \textbf{6117} & 0.64 & 0.58 \\  
		\cline{2-7}
		 D & 100           & 7318  & 7117 &  \textbf{6981} & 4.6 & 1.91 \\  
		\hline
		Layout  & 10 & 6248 & 6157 & \textbf{6085} & 2.60 & 1.17 \\  
		\cline{2-7}
		 E & 100 & 6847  & 7644 & \textbf{6774} & 1.06 & 11.38  \\  
		\hline
	\end{tabular}
	}
	\caption{ TTD time for $500$ tasks in seconds. We assume the robot moves $1$ unit in $1$ sec. We show results for Direct navigation scheme, which assumes no obstacles, and for A* navigation scheme on various layouts of grid size $60 \times 60$. Our method based on RL and decentralized navigation results in improving the overall performance (task completion time) over prior methods by $10.7\%$.	\vspace{-6mm}}
	 	\label{tab:simple_time}
\end{table}

\subsection{Navigation  with Multi-Agent Collision Avoidance}

\begin{figure}[t]
	\begin{subfigure}{.22\textwidth}
		\includegraphics[trim={0cm 0cm 0cm 0cm},scale=0.25]{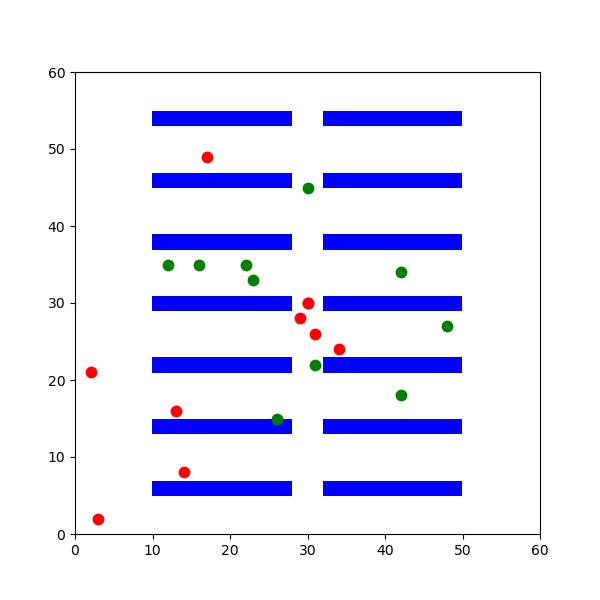}
		\vspace{-1mm}
		\caption{t = 0 sec.}
		\vspace{-1mm}
	\end{subfigure}%
	\begin{subfigure}{.22\textwidth}
		\includegraphics[trim={0cm 0cm 0cm 0cm},scale=0.25]{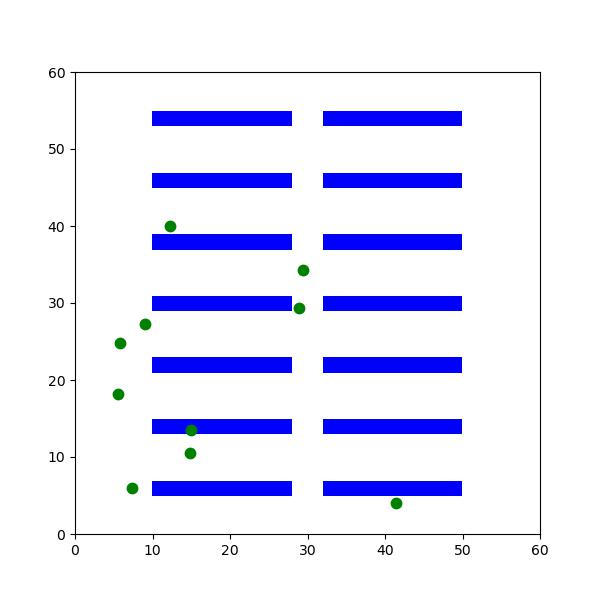}
		\caption{t = 42.5 sec.}
		\vspace{-1mm}
	\end{subfigure}%
	\hfill
	\begin{subfigure}{.22\textwidth}
		\includegraphics[trim={0cm 0cm 0cm 0cm},scale=0.25]{Figures/t_0.jpeg}
		\caption{t = 0 sec.}
	\end{subfigure}%
	\begin{subfigure}{.22\textwidth}
		\includegraphics[trim={0cm 0cm 0cm 0cm},scale=0.25]{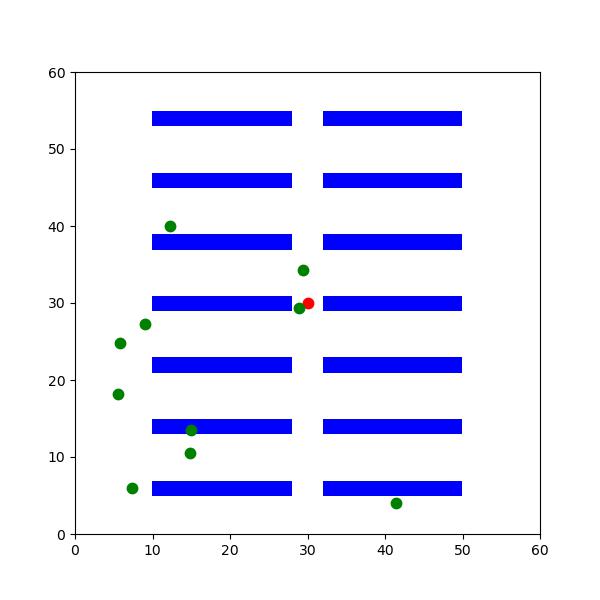}
		\caption{t = 42.5 sec.}
	\end{subfigure}
	\caption{We  compare the performance of DC-MRTA (top) with MPDM (bottom)   for a set of 10  agents performing  10  tasks. Agents are shown in green. The goal locations computed by the task allocation scheme are shown in red, which correspond to the destination positions for the agents. The initial configuration at $t=0$ is the same. At  $42.5$ seconds we observe that 10 tasks have been allocated with our method (top),  whereas the last task with MPDM is still unallocated, as shown by the red position.}
	\label{fig:visal_results}
	\vspace{-8mm}
\end{figure}

We evaluate the performance of DC-MRTA on all the layouts in Fig. \ref{fig:layout_other} and compare the performance for $10$ and $100$ robots summarized in \ref{tab:comp_nav}. We train on A* navigation algorithm and test on A*+ORCA (explained in section \ref{Astar_ORCA}). 
The task queue length is $10$ for all cases. As shown in Table \ref{tab:comp_nav}, DC-MRTA outperforms greedy (MPDM) as well as regret-based (RBTS) baselines for all layouts and navigation scenarios. We also see that for 100 agents improvement is lesser as compared to 10 agents. This is because crowding situation occurs more with 100 agents and our decentralised navigation scheme has to spend time to avoid obstacles. Thus it deviates from the time duration taught during training. But still it performs better than our baselines. We show visualisation of unallocated tasks left after 42.5 seconds for our method and MPDM baseline in Fig. \ref{fig:visal_results}.

\begin{table}[h]
	\centering
	\resizebox{1.0\columnwidth}{!}{%
	\begin{tabular}{|c|c|c|c|c|c|c|} 
		\hline
		\multicolumn{1}{|l|}{}             & Robot No. & MPDM  & RBTS & DC-MRTA & \% Imp. MPDM & \% Imp. RBTS \\
		\hline
		Layout  & 10 & 6499 & 6602 & \textbf{6251} & 3.82 & 5.32 \\  
		\cline{2-7}
		 A & 100 & 10616 & 10699 & \textbf{10339} & 2.61 &  3.36 \\  
		\hline
		Layout  & 10            & 5969 & 6526 & \textbf{5562} & 6.82 & 14.77 \\  
		\cline{2-7}
		 B & 100          & 9156 & 8957 & \textbf{8876} & 3.06 & 0.90 \\  
		\hline
		Layout  & 10            & 6566 & 6661 & \textbf{6084} & 7.34  & 8.66 \\  
		\cline{2-7}
		 C & 100           & 9709 & 9610 & \textbf{9590} & 1.23  & 0.21   \\  
		\hline
		Layout  & 10  & 5710 & 5845 & \textbf{5642} & 1.20  &  3.47  \\  
		\cline{2-7}
		 D & 100   & 8969 & 8924 & \textbf{8608} & 4.02 & 3.54  \\  
		\hline
		Layout  & 10 & 5983 &  5852 & \textbf{5538} & 7.44 & 5.37  \\  
		\cline{2-7}
		 E & 100 & 11224 & 10810 & \textbf{10423} & 7.14 & 3.58 \\  
		\hline
	\end{tabular}
	}
	\caption{ TTD time for $500$ tasks with $100$ robots  and $100$ tasks (in seconds) for 100 robots for A*+ORCA navigation scheme, respectively. The RL model is trained on only A* navigation scheme for various layouts and tested on A*+ORCA. We observe upto $14$\% improvement in the total time using our approach.} 
	 	\label{tab:comp_nav}
	 \vspace{-6mm}
\end{table}

\vspace{-3mm}
\subsection{Collision Values}
\vspace{-0.5mm}
We show number of inter-agent collisions with 10 agents  using A* and A*+ORCA navigation methods.

\begin{table}[H]
	\centering
		\resizebox{0.5\columnwidth}{!}{%
	\begin{tabular}{|c|c|c|c|c|c|}
		\hline
		Layout  & A & B & C &  D 	& E	\\
		\hline
		A*  & 5 & 6 & 2 & 4 & 2		\\
		\hline
		A*+ORCA & 2 & 0 & 0 & 2 & 2	\\
		\hline
	\end{tabular}}
	\caption{ This table highlights the  number of inter-agents collisions with two different navigation methods:  A* and A*+ORCA for 10 agents simulation with 10 tasks. Our RL model can be easily combined with A* + ORCA at testing time and results in 40\% reduction in number of collisions.}
		\label{tab:collision}
\end{table}

\vspace{-6mm}
\subsection{Scalability}\label{scalability}

We show that DC-MRTA is scalable by performing experiments with up to $1000$ robots. For direct navigation, we used a $300\times300$ grid layout with no obstacles. For A* we used a $256 \times 256$ grid size of layout C. The results are summarized in Table \ref{tab:scalable}. We note that DC-MRTA outperforms the baselines even for higher number of robots and tasks.

\begin{table}[H]
	\centering
		\resizebox{0.9\columnwidth}{!}{%
	\begin{tabular}{|c|c|c|c|c|c|}
		\hline
		Grid Size & MPDM  & RBTS & DC-MRTA & MPDM (\%) &  RBTS (\%) 
		\\
		\hline
		Direct & 695816 & 681236 & \textbf{674089} & 3.12 & 1.04 
		\\
		\hline
		A* on Layout B & 413406 & 423368 & \textbf{403112} & 2.49 & 4.78  
		\\
		\hline
	\end{tabular}
		}	\caption{ This table presents TTD in sec for $5000$ tasks and $1000$ robots for direct and A* navigation scheme. We assume the robot moves $1$ unit in $1$ sec. Thus, our approach is scalable and results in improved performance.}
		\label{tab:scalable}
\end{table}

\vspace{-6mm}
\section{Conclusion, Limitations, and Future work}
We present a novel approach for the problem of joint task allocation and decentralized multi-robot navigation in a complex warehouse environments. We propose a novel two level coupled approach approach, where   lower level collision free decentralized navigation is combine with high level RL-based task allocation. The reward functions for the RL method are defined based on the feedback from the navigation method. We have evaluated our method on complex environments and obtain up to $14\%$ improvement over prior methods in terms of task completion time. Our RL model can be combined with any decentralized method, and we highlight its benefits in terms of collision avoidance using ORCA.

Our approach has some limitations. The decentralized schemes may not always result in optimal trajectories, or collision-free paths. Our formulation of coupled rewards can be further improved based on other feedback that takes into account agent density and congestion. Our approach is limited to homogeneous agents with simple dynamics and assume that all agents have precise information about the obstacles in the scene. There are many avenues for future work. We would like to evaluate other rewards functions that can help reduce the congestion or challenging configurations at runtime. We would like to further evaluate the performance, where the layout changes or there are dynamic, human-like obstacles. Another interesting future line of work we aim to explore is limited communication. This means that the agents in decentralised system even though cannot communicate with the central server, it can communicate with other surrounding agents. They will then take actions in coordination with the neighboring agents.  

{\small
	\bibliographystyle{IEEEtran}
	\bibliography{sample,ijcai22}

\begin{thebibliography}{10}
\providecommand{\url}[1]{#1}
\csname url@rmstyle\endcsname
\providecommand{\newblock}{\relax}
\providecommand{\bibinfo}[2]{#2}
\providecommand\BIBentrySTDinterwordspacing{\spaceskip=0pt\relax}
\providecommand\BIBentryALTinterwordstretchfactor{4}
\providecommand\BIBentryALTinterwordspacing{\spaceskip=\fontdimen2\font plus
\BIBentryALTinterwordstretchfactor\fontdimen3\font minus
  \fontdimen4\font\relax}
\providecommand\BIBforeignlanguage[2]{{%
\expandafter\ifx\csname l@#1\endcsname\relax
\typeout{** WARNING: IEEEtran.bst: No hyphenation pattern has been}%
\typeout{** loaded for the language `#1'. Using the pattern for}%
\typeout{** the default language instead.}%
\else
\language=\csname l@#1\endcsname
\fi
#2}}

\bibitem{planetary}
C.~R. Weisbin and G.~Rodriguez, ``Nasa robotics research for planetary surface
  exploration,'' \emph{IEEE Robotics \& Automation Magazine}, vol.~7, no.~4,
  pp. 25--34, 2000.

\bibitem{searchandrescue}
J.~S. Jennings, G.~Whelan, and W.~F. Evans, ``Cooperative search and rescue
  with a team of mobile robots,'' in \emph{1997 8th International Conference on
  Advanced Robotics. Proceedings. ICAR'97}.\hskip 1em plus 0.5em minus
  0.4em\relax IEEE, 1997, pp. 193--200.

\bibitem{automated_manufacturing}
J.~Tilley, ``Automation, robotics, and the factory of the future,''
  \emph{McKinsey. https://www. mckinsey.
  com/business-functions/operations/our-insights/automation-robotics-and-the-factory-of-the-future},
  2017.

\bibitem{pick_up_and_delivery}
M.~Liu, H.~Ma, J.~Li, and S.~Koenig, ``Task and path planning for multi-agent
  pickup and delivery,'' in \emph{Proceedings of the International Joint
  Conference on Autonomous Agents and Multiagent Systems (AAMAS)}, 2019.

\bibitem{warehouse_comb}
F.~Xue, H.~Tang, Q.~Su, and T.~Li, ``Task allocation of intelligent warehouse
  picking system based on multi-robot coalition,'' \emph{KSII Transactions on
  Internet and Information Systems}, vol.~13, no.~7, 2019.

\bibitem{mapf_decoupled}
H.~Ma, J.~Li, T.~K.~S. Kumar, and S.~Koenig, ``Lifelong multi-agent path
  finding for online pickup and delivery tasks,'' 2017.

\bibitem{market_effi}
A.~Khamis, A.~Elmogy, and F.~Karray, ``Complex task allocation in mobile
  surveillance systems,'' \emph{Journal of Intelligent and Robotic Systems},
  vol.~64, pp. 33--55, 10 2011.

\bibitem{market_robust}
M.~B. Dias and A.~Stentz, ``A free market architecture for distributed control
  of a multirobot system,'' 2000.

\bibitem{astar}
M.~Goldenberg, A.~Felner, R.~Stern, G.~Sharon, N.~Sturtevant, R.~C. Holte, and
  J.~Schaeffer, ``Enhanced partial expansion a*,'' \emph{J. Artif. Int. Res.},
  vol.~50, no.~1, p. 141–187, May 2014.

\bibitem{orca}
J.~van~den Berg, S.~J. Guy, M.~Lin, and D.~Manocha, ``Reciprocal n-body
  collision avoidance,'' in \emph{Robotics Research}.\hskip 1em plus 0.5em
  minus 0.4em\relax Berlin, Heidelberg: Springer Berlin Heidelberg, 2011, pp.
  3--19.

\bibitem{sipp}
M.~Phillips and M.~Likhachev, ``Sipp: Safe interval path planning for dynamic
  environments,'' in \emph{IEEE ICRA}.\hskip 1em plus 0.5em minus 0.4em\relax
  IEEE, 2011, pp. 5628--5635.

\bibitem{cbs}
\BIBentryALTinterwordspacing
G.~Sharon, R.~Stern, A.~Felner, and N.~R. Sturtevant, ``Conflict-based search
  for optimal multi-agent pathfinding,'' \emph{Artificial Intelligence}, vol.
  219, pp. 40--66, 2015. [Online]. Available:
  \url{https://www.sciencedirect.com/science/article/pii/S0004370214001386}
\BIBentrySTDinterwordspacing

\bibitem{vrvo}
S.~H. Arul and D.~Manocha, ``V-rvo: Decentralized multi-agent collision
  avoidance using voronoi diagrams and reciprocal velocity obstacles,'' 2021.

\bibitem{Luna2011PushAS}
R.~Luna and K.~E. Bekris, ``Push and swap: Fast cooperative path-finding with
  completeness guarantees,'' in \emph{IJCAI}, 2011.

\bibitem{MILP}
T.~Schouwenaars, B.~De~Moor, E.~Feron, and J.~How, ``Mixed integer programming
  for multi-vehicle path planning,'' in \emph{2001 European Control Conference
  (ECC)}, 2001, pp. 2603--2608.

\bibitem{MIQP}
D.~Mellinger, A.~Kushleyev, and V.~Kumar, ``Mixed-integer quadratic program
  trajectory generation for heterogeneous quadrotor teams,'' in \emph{2012 IEEE
  International Conference on Robotics and Automation}, 2012, pp. 477--483.

\bibitem{velagapudi2010decentralized}
P.~Velagapudi, K.~Sycara, and P.~Scerri, ``Decentralized prioritized planning
  in large multirobot teams,'' in \emph{2010 IEEE/RSJ International Conference
  on Intelligent Robots and Systems}.\hskip 1em plus 0.5em minus 0.4em\relax
  IEEE, 2010, pp. 4603--4609.

\bibitem{cohen2019optimal}
L.~Cohen, T.~Uras, T.~S. Kumar, and S.~Koenig, ``Optimal and bounded-suboptimal
  multi-agent motion planning,'' in \emph{Twelfth Annual Symposium on
  Combinatorial Search}, 2019.

\bibitem{wilkie2009generalized}
D.~Wilkie, J.~Van Den~Berg, and D.~Manocha, ``Generalized velocity obstacles,''
  in \emph{2009 IEEE/RSJ International Conference on Intelligent Robots and
  Systems}.\hskip 1em plus 0.5em minus 0.4em\relax IEEE, 2009, pp. 5573--5578.

\bibitem{alonso2013optimal}
J.~Alonso-Mora, A.~Breitenmoser, M.~Rufli, P.~Beardsley, and R.~Siegwart,
  ``Optimal reciprocal collision avoidance for multiple non-holonomic robots,''
  in \emph{Distributed autonomous robotic systems}.\hskip 1em plus 0.5em minus
  0.4em\relax Springer, 2013, pp. 203--216.

\bibitem{usc}
Z.~Chen, J.~Alonso-Mora, X.~Bai, D.~D. Harabor, and P.~J. Stuckey, ``Integrated
  task assignment and path planning for capacitated multi-agent pickup and
  delivery,'' \emph{IEEE Robotics and Automation Letters}, vol.~6, no.~3, pp.
  5816--5823, 2021.

\bibitem{VO}
P.~Fiorini and Z.~Shiller, ``Motion planning in dynamic environments using
  velocity obstacles,'' \emph{The International Journal of Robotics Research},
  vol.~17, no.~7, pp. 760--772, 1998.

\bibitem{RVO}
J.~van~den Berg, M.~Lin, and D.~Manocha, ``Reciprocal velocity obstacles for
  real-time multi-agent navigation,'' in \emph{2008 IEEE International
  Conference on Robotics and Automation}, 2008, pp. 1928--1935.

\bibitem{AVO}
J.~Van Den~Berg, J.~Snape, S.~J. Guy, and D.~Manocha, ``Reciprocal collision
  avoidance with acceleration-velocity obstacles,'' in \emph{2011 IEEE
  International Conference on Robotics and Automation}.\hskip 1em plus 0.5em
  minus 0.4em\relax IEEE, 2011, pp. 3475--3482.

\bibitem{NH-ORCA}
J.~Alonso-Mora, A.~Breitenmoser, M.~Rufli, P.~Beardsley, and R.~Siegwart,
  \emph{Optimal Reciprocal Collision Avoidance for Multiple Non-Holonomic
  Robots}.\hskip 1em plus 0.5em minus 0.4em\relax Berlin, Heidelberg: Springer
  Berlin Heidelberg, 2013, pp. 203--216.

\bibitem{BVC}
D.~Zhou, Z.~Wang, S.~Bandyopadhyay, and M.~Schwager, ``Fast, on-line collision
  avoidance for dynamic vehicles using buffered voronoi cells,'' \emph{IEEE
  Robotics and Automation Letters}, vol.~2, no.~2, pp. 1047--1054, 2017.

\bibitem{fan2020distributed}
T.~Fan, P.~Long, W.~Liu, and J.~Pan, ``Distributed multi-robot collision
  avoidance via deep reinforcement learning for navigation in complex
  scenarios,'' \emph{The International Journal of Robotics Research}, 2020.

\bibitem{GA3C_CADRL}
S.~H. Semnani, H.~Liu, M.~Everett, A.~de~Ruiter, and J.~P. How, ``Multi-agent
  motion planning for dense and dynamic environments via deep reinforcement
  learning,'' \emph{IEEE Robotics and Automation Letters}, vol.~5, no.~2, pp.
  3221--3226, 2020.

\bibitem{tan2020deepmnavigate}
Q.~Tan, T.~Fan, J.~Pan, and D.~Manocha, ``Deepmnavigate: Deep reinforced
  multi-robot navigation unifying local \& global collision avoidance,'' in
  \emph{2020 IEEE/RSJ International Conference on Intelligent Robots and
  Systems (IROS)}.\hskip 1em plus 0.5em minus 0.4em\relax IEEE, 2020, pp.
  6952--6959.

\bibitem{yousefikhoshbakht2013modification}
M.~Yousefikhoshbakht, F.~Didehvar, and F.~Rahmati, ``Modification of the ant
  colony optimization for solving the multiple traveling salesman problem,''
  \emph{Romanian Journal of Information Science and Technology}, vol.~16,
  no.~1, pp. 65--80, 2013.

\bibitem{6721865}
J.~Li, Q.~Sun, M.~Zhou, and X.~Dai, ``A new multiple traveling salesman problem
  and its genetic algorithm-based solution,'' in \emph{2013 IEEE International
  Conference on Systems, Man, and Cybernetics}, 2013, pp. 627--632.

\bibitem{taxonomy}
\BIBentryALTinterwordspacing
B.~P. Gerkey and M.~J. Matarić, ``A formal analysis and taxonomy of task
  allocation in multi-robot systems,'' \emph{The International Journal of
  Robotics Research}, vol.~23, no.~9, pp. 939--954, 2004. [Online]. Available:
  \url{https://doi.org/10.1177/0278364904045564}
\BIBentrySTDinterwordspacing

\bibitem{Khamis2015}
A.~Khamis, A.~Hussein, and A.~Elmogy, \emph{Multi-robot Task Allocation: A
  Review of the State-of-the-Art}.\hskip 1em plus 0.5em minus 0.4em\relax Cham:
  Springer International Publishing, 2015, pp. 31--51.

\bibitem{Tkach2020}
I.~Tkach and Y.~Edan, \emph{Multi-agent Task Allocation}.\hskip 1em plus 0.5em
  minus 0.4em\relax Cham: Springer International Publishing, 2020, pp. 9--14.

\bibitem{warehouse_hetro_robot}
\BIBentryALTinterwordspacing
N.~Baras, A.~Chatzisavvas, D.~Ziouzios, and M.~Dasygenis, ``Improving automatic
  warehouse throughput by optimizing task allocation and validating the
  algorithm in a developed simulation tool,'' \emph{Automation}, vol.~2, no.~3,
  pp. 116--126, 2021. [Online]. Available:
  \url{https://www.mdpi.com/2673-4052/2/3/7}
\BIBentrySTDinterwordspacing

\bibitem{badreldin2013comparative}
M.~Badreldin, A.~Hussein, and A.~Khamis, ``A comparative study between
  optimization and market-based approaches to multi-robot task allocation.''
  \emph{Advances in Artificial Intelligence (16877470)}, 2013.

\bibitem{brunet2008consensus}
L.~Brunet, H.-L. Choi, and J.~How, ``Consensus-based auction approaches for
  decentralized task assignment,'' in \emph{AIAA guidance, navigation and
  control conference and exhibit}, 2008, p. 6839.

\bibitem{choudhury2021dynamic}
S.~Choudhury, J.~K. Gupta, M.~J. Kochenderfer, D.~Sadigh, and J.~Bohg,
  ``Dynamic multi-robot task allocation under uncertainty and temporal
  constraints,'' \emph{Autonomous Robots}, pp. 1--17, 2021.

\bibitem{CA-CBBA}
S.~Raja, G.~Habibi, and J.~P. How, ``Communication-aware consensus-based
  decentralized task allocation in communication constrained environments,''
  \emph{IEEE Access}, vol.~10, pp. 19\,753--19\,767, 2022.

\bibitem{ames_decTask}
Y.~Chen, U.~Rosolia, and A.~D. Ames, ``Decentralized task and path planning for
  multi-robot systems,'' \emph{IEEE Robotics and Automation Letters}, vol.~6,
  no.~3, pp. 4337--4344, 2021.

\bibitem{ghassemi2018decentralized}
P.~Ghassemi and S.~Chowdhury, ``Decentralized task allocation in multi-robot
  systems via bipartite graph matching augmented with fuzzy clustering,'' in
  \emph{International design engineering technical conferences and computers
  and information in engineering conference}, vol. 51753.\hskip 1em plus 0.5em
  minus 0.4em\relax American Society of Mechanical Engineers, 2018, p.
  V02AT03A014.

\bibitem{liu2013optimal}
L.~Liu and D.~A. Shell, ``Optimal market-based multi-robot task allocation via
  strategic pricing.'' in \emph{Robotics: Science and Systems}, vol.~9, no.~1,
  2013, pp. 33--40.

\bibitem{task_survey}
G.~A. Korsah, A.~Stentz, and M.~B. Dias, ``A comprehensive taxonomy for
  multi-robot task allocation,'' \emph{The International Journal of Robotics
  Research}, vol.~32, no.~12, pp. 1495--1512, 2013.

\bibitem{DBLP:journals/corr/abs-1909-01150}
\BIBentryALTinterwordspacing
L.~Wang, Q.~Cai, Z.~Yang, and Z.~Wang, ``Neural policy gradient methods: Global
  optimality and rates of convergence,'' \emph{CoRR}, vol. abs/1909.01150,
  2019. [Online]. Available: \url{http://arxiv.org/abs/1909.01150}
\BIBentrySTDinterwordspacing

\bibitem{attention}
A.~Vaswani, N.~Shazeer, N.~Parmar, J.~Uszkoreit, L.~Jones, A.~N. Gomez,
  {\L}.~Kaiser, and I.~Polosukhin, ``Attention is all you need,'' in
  \emph{NeuRIPS}, 2017.

\bibitem{taxi_comb_opti}
L.~Zhang, T.~Hu, Y.~Min, G.~Wu, J.~Zhang, P.~Feng, P.~Gong, and J.~Ye, ``A taxi
  order dispatch model based on combinatorial optimization,'' in
  \emph{Proceedings of the 23rd ACM SIGKDD International Conference on
  Knowledge Discovery and Data Mining}, ser. KDD '17.\hskip 1em plus 0.5em
  minus 0.4em\relax New York, NY, USA: Association for Computing Machinery,
  2017, p. 2151–2159.

\bibitem{rbts_1}
F.~A. Tillman and T.~M. Cain, ``An upperbound algorithm for the single and
  multiple terminal delivery problem,'' \emph{Management Science}, vol.~18,
  no.~11, pp. 664--682, 1972.

\bibitem{rbts_2}
S.~K.~X. Zheng, C.~Tovey, R.~Borie, P.~Kilby, V.~Markakis, and P.~Keskinocak,
  ``Agent coordination with regret clearing,'' in \emph{AAAI}, 2008.

\bibitem{mapd_review}
R.~Stern, N.~Sturtevant, A.~Felner, S.~Koenig, H.~Ma, T.~Walker, J.~Li,
  D.~Atzmon, L.~Cohen, T.~Kumar, E.~Boyarski, and R.~Barták, ``Multi-agent
  pathfinding: Definitions, variants, and benchmarks,'' 06 2019.

\bibitem{task_gen}
L.~Cohen, T.~Uras, and S.~Koenig, ``Feasibility study: Using highways for
  bounded-suboptimal multi-agent path finding,'' in \emph{Eighth Annual
  Symposium on Combinatorial Search}, 2015.

\bibitem{task_gen2}
H.~Ma, D.~Harabor, P.~J. Stuckey, J.~Li, and S.~Koenig, ``Searching with
  consistent prioritization for multi-agent path finding,'' in \emph{AAAI},
  vol.~33, no.~01, 2019, pp. 7643--7650.

\end{thebibliography}
}

\end{document}